# Chapter 42

# Bilinear Fuzzy Genetic algorithm and its application on the optimum design of steel structures with semi-rigid connections

Salar Farahmand-Tabar[1*] · Payam Ashtari[1]
[1]Department of Civil Engineering Eng., Faculty of Engineering, University of Zanjan, Zanjan, Iran,
*farahmandsalar@znu.ac.ir; farahmandsalar@gmail.com
ashtari@znu.ac.ir

**Abstract.** An improved bilinear fuzzy genetic algorithm (BFGA) is introduced in this chapter for the design optimization of steel structures with semi-rigid connections. Semi-rigid connections provide a compromise between the stiffness of fully rigid connections and the flexibility of fully pinned connections. However, designing such structures is challenging due to the non-linear behavior of semi-rigid connections. The BFGA is a robust optimization method that combines the strengths of fuzzy logic and genetic algorithm to handle the complexity and uncertainties of structural design problems. The BFGA, compared to standard GA, demonstrated to generate high-quality solutions in a reasonable time. The application of the BFGA is demonstrated through the optimization of steel structures with semi-rigid connections, considering the weight, and performance criteria. The results show that the proposed BFGA is capable of finding optimal designs that satisfy all the design requirements and constraints. The proposed approach provides a promising solution for the optimization of complex structures with non-linear behavior.

## 1. Introduction

Semi-rigid connections are defined as connections that exhibit partial moment-resisting behavior, meaning that they allow some degree of rotation between the two connected members, but also provide some resistance to lateral movement. These connections typically include elements such as bolts, welding, or other fasteners that provide a certain level of stiffness while allowing some rotational flexibility. Semi-rigid connections are commonly used in building frames, bridges, and industrial structures where rotational flexibility is required but excessive movement must be avoided. For example, in building frames, semi-rigid connections are used to connect beams to columns, allowing the structure to resist lateral loads while also providing some flexibility to accommodate movement due to wind or seismic activity.

Semi-rigid connections offer several advantages over fully rigid or fully flexible connections. One major advantage is that they can provide a higher level of ductility in the structure, which is the ability of the structure to deform without breaking during extreme events such as earthquakes. This is because semi-rigid connections can accommodate some level of movement without compromising the overall structural integrity. Another advantage of semi-rigid connections is that they can be more cost-effective than fully rigid connections. This is because they require less material and can be easier to fabricate and install than fully rigid connections. Additionally, semi-rigid connections can provide some level of energy dissipation, which can help to reduce the overall dynamic response of the structure to external loads. However, they can be more complex to design and analyze than fully rigid or fully flexible connections because the behavior of semi-rigid connections can be more difficult to predict, and the design must take into account the interaction between the connected members, the connection details, and the loading conditions.

Structural optimization is a crucial area of research in engineering, as it allows for the design of structures that are both efficient and cost-effective [1]. The use of semi-rigid connections in structures is an important consideration in this process, as these connections provide some degree of rotational flexibility while also maintaining a level of stiffness. Despite significant progress in understanding the behavior and design of semi-rigid connections, there remains a gap in the literature when it comes to optimizing structures with these connections.

In particular, previous studies have mostly focused on the behavior and design of either fully rigid or fully flexible connections. Several methods have been proposed for the optimization of structures with semi-rigid connections, ranging from simplified analytical approaches to more complex numerical simulations. These methods typically take into account the behavior of the connections and their interaction with other structural elements, as well as the overall stiffness and load-bearing capacity of the structure.

Structural elements such as castellated beams with semi-rigid connection were considered as optimization problems [2, 3]. Grid structures with semi-rigid connections were also subjected to the stiffness and shape optimization [4, 5]. Semi-rigid connections can be optimally designed and constructed in different types of frames structures such timber buildings [6], and reinforced concrete structures, [7, 8]. Research on the optimum application of semi-rigid connections in steel structures has been a topic of interest in the field of structural engineering, as demonstrated by several recent studies. Design Optimization of semi-rigid steel frames have been carried out using various metaheuristic algorithms such as Genetic Algorithms [9-12], generalized reduced gradient algorithm [13], Bees and Genetic Algorithms [14], Harmony Search and Particle Swarm Optimization [15, 16], big bang-big brunch [17], and hybrid Harmony Search and Big Bang-Big Crunch [18]. Semi-rigid connections are also used optimally in composite frames [19-21]. There are some research papers that consider the behavioral parameters and analysis types such as stability [22], second-order analysis [23, 24], CO2 emission [25], Virtual work [26, 27], and reliability [28] in the design optimization of frames with semi-rigid connections. By the advances in computational intelligence, surrogate models (metamodels) can be included in the optimum design process of these structures [29].

Fuzzified optimization is a technique that combines fuzzy logic with optimization methods to improve the accuracy and robustness of the optimization process. In the context of optimizing structures with semi-rigid connections, adding fuzziness to the algorithms such as Genetic Algorithm (GA) can help to better capture the uncertain and imprecise nature of the design variables and constraints [30-37]. This is particularly important in situations where the behavior of the semi-rigid connections may not be fully understood or where there is significant variability in the loading and environmental conditions.

The current chapter proposes a bilinear fuzzy method for design optimization of structures with semi-rigid connections. This method takes into account the flexibility of the connections, while also considering the overall stiffness of the structure. By doing so, it aims to find the optimal balance between flexibility and stiffness, leading to more efficient and cost-effective designs. Through a series of numerical simulations and case studies, the effectiveness of the proposed method is demonstrated. It is shown how it can lead to significant improvements in the performance and cost-effectiveness of structures with semi-rigid connections.

## 2. Bilinear Fuzzy Genetic Algorithm

Numerous optimization applications have made extensive use of the Genetic Algorithm (GA) and its fuzzy variant (Fuzzy-GA) [36-39]. The performance of the fuzzy GA can be influenced by the membership function used through the fuzzification process. Here, the bi-linear membership function is utilized in Fuzzy-GA for both constraints and objective function to accelerate its performance.

According to crisp logic, a phenomenon can either be true or false. The application of crisp logic will complicate the circumstances for structural optimal design since the problem is affected by a significant level of ambiguity and imprecision. Fuzzy logic takes into account the transition between exclusion and inclusion using a membership function, $\mu_Y$, which converts each member of set $X$ into a fuzzy set $Y$ by doing the following:

$$\begin{cases} \mu_Y(x) & \in & [0,1] \\ x & \in & X \end{cases} \tag{1}$$

The value of membership in crisp logic for structural element $i$ is equal to 1 if $\sigma_i < \sigma^u$ and equal to zero if the constraint is not fulfilled. However, the membership value ($\alpha$) is altered from 0 to 1 for fuzzy logic, where $\alpha = 1$ denotes that the constraint is fulfilled and $\alpha = 0$ denotes that the constraint has been violated relative to the specified tolerance. Consequently, $0 < \alpha < 1$ denotes that the constraint has been met to the specified amount. When using fuzzy optimization with the membership function, both inequality constraints and the objective function are fuzzified.

$$\mu_{\bar{D}}(x) = \mu_F(x) \cap \left\{ \bigcap_{i=1}^{m} \mu_{g_i}(x) \right\} \tag{2}$$

where, $\mu_{g_i}(x)$ and $\mu_F(x)$ denote the membership functions for the $i^{th}$ inequality constraint and the objective function, respectively. The $x_{opt}$ is achieved from the region as:

$$\mu_{\bar{D}}(x_{opt}) = \max \mu_{\bar{D}}(x), \qquad \mu_{\bar{D}}(x) = \min\left\{ \mu_F(x), \min_{i=1,\ldots,m} \mu_{g_i}(x) \right\} \tag{3}$$

The membership functions related to the objective function and constraints (Fig. 1) are intersected to provide a fuzzy domain. The optimization problem can be expressed using a general satisfaction parameter (λ) as follows:

maximize $\lambda$ subject to

$$\begin{aligned} &\lambda \leq \mu_F(x) \\ &\lambda \leq \mu_{g_i}(x); \quad i = 1, \dots, m \\ &0 \leq \lambda \leq 1 \end{aligned} \tag{4}$$

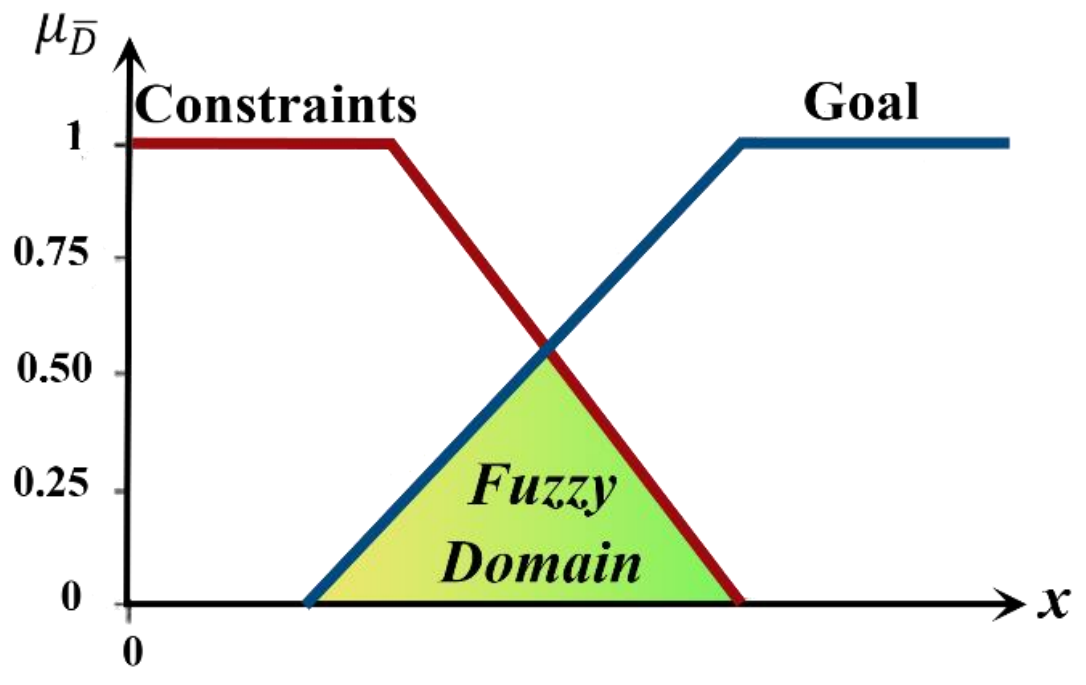

**Fig. 1** Intersection of membership functions

An optimization approach is suggested using a formulation that combines fuzzy logic with enhanced Lagrangian penalty functions considering the membership parameters of $\alpha$ and $\beta$ for the objective function and $\gamma$ and $\omega$ for the constraints.

To minimize

$$\begin{aligned} \varphi(x,y) = -S_f\lambda + \frac{1}{2}\left\{\alpha\left[\frac{\lambda}{\mu_F} - 1 + \beta\right]^2\right\} + \frac{1}{2}\left\{\sum_{i=1}^{noe}\gamma_i\left[\frac{\lambda}{\mu_{\sigma i}{}^a} - 1 + \omega_i\right]^2\right\} \\ + \frac{1}{2}\left\{\sum_{j=1}^{nod}\gamma_j\left[\frac{\lambda}{\mu_{\delta j}} - 1 + \omega_j\right]^2\right\} + \frac{1}{2}\left\{\sum_{k=1}^{nm}\gamma_k\left[\frac{\lambda}{\mu_{gk}} - 1 + \omega_k\right]^2\right\} \end{aligned} \tag{5}$$

It is intended to maximize the penalized $\lambda$ ($S_f\lambda$) while taking into account the constraints. In structural optimization problems, stress ($\sigma$), displacement ($\delta$), and constructability ($g$) are typically the kinds of constraints expressed as follows using the membership function ($\mu$):

$$\begin{aligned} &\lambda \leq \mu_{\sigma_i}^{ub}(x); \quad \lambda \leq \mu_{\sigma_i}^{lb}(x) \qquad i = 1, \dots, \textit{total number of members (nm)} \\ &\lambda \leq \mu_{\delta_j}(x); \qquad j = 1, \dots, \textit{number of constrained DOFs (nd)} \\ &\lambda \leq \mu_{g_k}(x); \qquad k = 1, \dots, \textit{total number of diagrid units (nu)} \end{aligned} \tag{6}$$

All constraints can be expressed as $min(\mu_F(x), \min_{i=1,\dots,nm} \mu_{\sigma_i}^a, \min_{j=1,\dots,nd} \mu_{\delta_j}, \min_{k=1,\dots,nu} \mu_{g_k})$ to form an unconstrained problem as:

$$\text{maximize } \varphi(x) = \lambda \tag{7}$$

First, each chromosome is given a design variable (x) via GA. Then, using structural analysis, membership values for the objective functions and constraints are determined. Using Eq (9), the fitness value of each chromosome is determined. It's crucial to pick the right membership function for the objective function to reach a solution. If the objective function's lower and upper bounds for minimization problems are assumed to be F' and F", then

$$\mu_F = \begin{cases} 1\,; & if \;\; F \le F^{'} \\ 0\,; & if \;\; F > F^{''} \end{cases} \tag{8}$$

The membership value ($F' <$ F $< F''$) can be used to implement the necessary membership function. The $F$" may be calculated using basic GA iterations or by using design experience. Even if the violated constraints emerge, $F$" can be lowered to a minimal value or a specific portion of the objective function to reach the value of the lower bound ($F'$). There might be several issues with the objective function's linear membership function. The possibility of an expedited convergence to a local optimum solution is raised if the fuzzy-GA technique is employed after the process of standard GA.

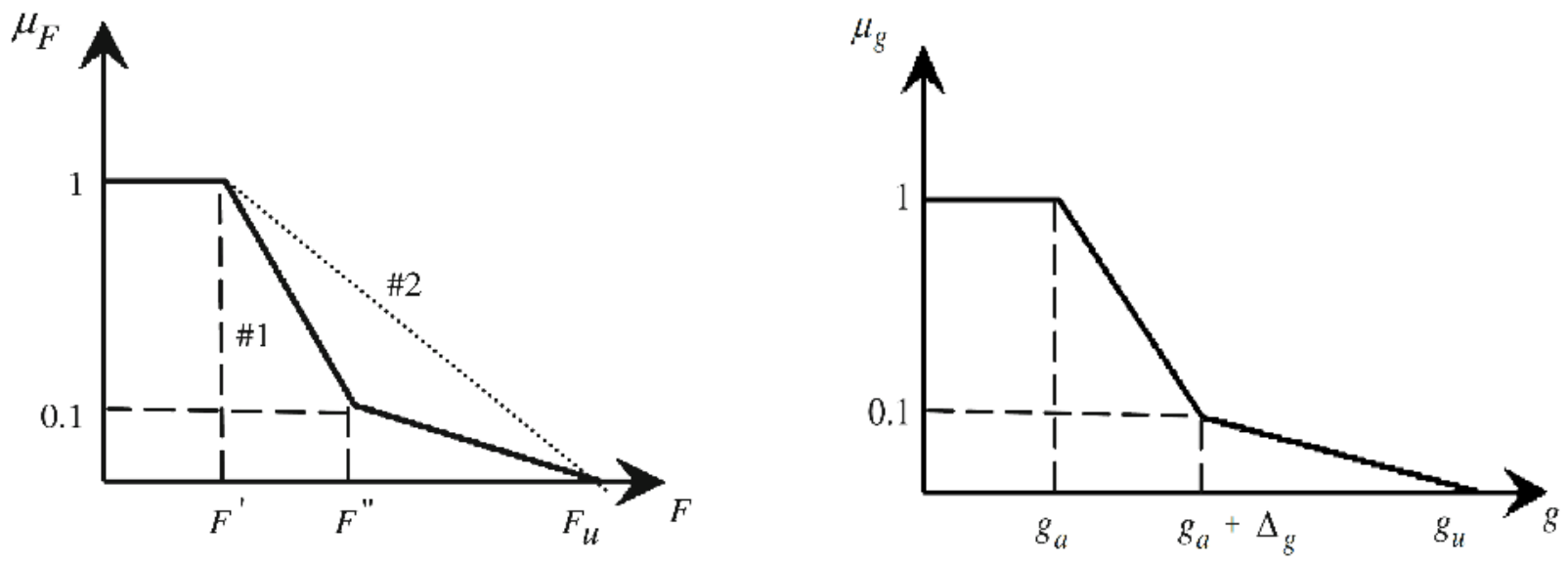

**a)** Objective function **b)** Constraint

**Fig. 2** Bi-linear membership functions [36]

However, In the initial iterations of the fuzzy-GA, we frequently have $F > F$" if $F$" is identical to the best solution, and the membership value ($\mu_F$) for the majority of chromosomes becomes zero, making it impossible to compare them with one another. Therefore, it is possible to accelerate the convergence by using a bi-linear membership function (Fig. 2.b). If membership function No. 2 is taken into account in Fig. 2(a), $F_u \le F$, the slope of the line decreases relative to line No. 1, indicating that there is less diversity in the fitness values of the chromosomes.

As a result, the possibility of the algorithm to be trapped in a locally optimal solution will rise, which will lead to a lower convergence rate. The value of $F''$ in Fig. 2(a) can either be the average of $F'$ and $F_u$ or the best value from earlier studies. The allowable values ($g_a$) of the constraints can be used to choose the membership function for them. The applied membership function for the constraints in the current study is displayed in Fig. 2(b), where $g_u$ is equal to $n \times g_a$ ($n > 1$) and $\Delta_g$ is the amount of the relaxation for the constraint.

**Algorithm 1**. The Pseudocode of GA with accelerating fuzzification

```
% Initialization
    Population Size = α;   Elitism Rate = β;   Mutation Rate = γ;
    Generate α random feasible solutions; Save them in Pop;
% Main Loop
  for i = 1: Max_Generations
% Fitness Evaluation
      Decoding chromosomes
      Fuzzification by membership function on constraints||objective function
      Fitness evaluation
% Selection
      Number of Elitisms ne = α.β
      Select the best ne solutions in Pop and save them in Pop1
% Crossover
      Number of Crossovers nc = (α − ne)/2
      if Chromosomes length for parents are unequal to those of children
          Apply length increment on parent chromosomes
      end
      for j = 1: nc
         Randomly select two solutions X_A and X_B from Pop
         Generate X_C and X_D by one-point crossover to X_A and X_B
         Save X_C and X_D to Pop2
      end
% Mutation
      for j = 1: nc
         Select a solution X_j from Pop2
         Mutate each bit of X_j under the rate γ and generate a new solution X_j'
         if X_j' is unfeasible; update X_j' with a feasible solution by repairing X_j'
         end
         update X_j with X_j' in Pop2
      end
% Updating
         Update Pop = Pop1 + Pop2
  end
% Final Solution
Return the best solution X in Pop
```

## 3. Computational modeling of semi-rigid connections

An applied moment $M$ causes a connection to rotate through an angle $\theta$, which is the angle between the original positions of the beam and column. One approach taken by researchers to model semi-rigid connections in steel frames is to use moment-rotation relationships achieved from experimental investigations. Fig. 3 illustrates various types of moment-rotation curves for these connections that have been derived by researchers. Semi-rigid beam connections can be modeled using rotational springs, as illustrated in Figs. 4 and 5, where $\theta_{rA}$ and $\theta_{rB}$ represent the relative spring rotations of both ends, and $K_A$ and $K_B$ are the corresponding spring stiffness.

$$K_A = \frac{K_A}{\theta_{rA}}, \quad K_B = \frac{K_B}{\theta_{rB}} \tag{9}$$

The stiffness matrix in global coordinates for beam member $(i)$ with semi-rigid end connections can be expressed as:

$$[K]_i = \begin{bmatrix} b_1 & & & & & \\ b_3 & b_2 & & & & \\ -b_4 & b_5 & b_6 & & & \\ -b_1 & -b_3 & b_4 & b_1 & & \\ -b_3 & -b_2 & -b_5 & b_3 & b_2 & \\ -b_7 & b_8 & b_9 & b_7 & -b_8 & b_{10} \end{bmatrix} \tag{10}$$

where,

$$b_1 = \frac{EA}{L}\cos^2\alpha + \frac{12}{K_R}\left(\frac{EI}{L^2}\right)^2\left(\frac{L}{EI}\frac{1}{K_A}\frac{1}{K_B}\right)\sin^2\theta$$

$$b_2 = \frac{EA}{L}\sin^2\alpha + \frac{12}{K_R}\left(\frac{EI}{L^2}\right)^2\left(\frac{L}{EI}\frac{1}{K_B}\frac{1}{K_A}\right)\cos^2\theta$$

$$b_3 = \frac{EA}{L}\sin\alpha\cos\alpha + \frac{12}{K_R}\left(\frac{EI}{L^2}\right)^2\left(\frac{L}{EI}\frac{1}{K_B}\frac{1}{K_A}\right)\sin\theta\cos\theta$$

$$b_4 = \frac{12}{K_R}\left(\frac{EI}{L}\right)^2\left(\frac{1}{2EI}\frac{1}{K_B}\right)\sin\alpha \qquad b_5 = \frac{12}{LK_R}\left(\frac{EI}{L^2}\right)^2\left(\frac{1}{2EI}\frac{1}{K_B}\right)\sin\alpha\cos\alpha$$

$$b_6 = \frac{12}{K_R}\left(\frac{EI}{L}\right)^2\left(\frac{1}{2EI}\frac{1}{K_B}\right) \qquad b_7 = \frac{12}{LK_R}\left(\frac{EI}{L}\right)^2\left(\frac{1}{2EI}\frac{1}{K_A}\right)\sin\alpha$$

$$b_8 = \frac{12}{LK_R}\left(\frac{EI}{L}\right)^2\left(\frac{1}{2EI}\frac{1}{K_A}\right)\cos\alpha \qquad b_9 = \frac{2EI}{LK_R}, \quad b_{10} = \frac{12}{K_R}\left(\frac{EI}{L}\right)^2\left(\frac{L}{3EI}\frac{1}{K_A}\right)$$

$$K_R = \left(1 + \frac{4EI}{LK_A}\right)\left(1 + \frac{4EI}{LK_B}\right) - \left(\frac{EI}{L}\right)^2\left(\frac{4}{K_AK_B}\right)$$

$$M_{FA} = \frac{M_{FA} + 6\alpha_B M_{FB}}{(1 + 4\alpha_A + 4\alpha_B + 12\alpha_A\alpha_B)} \qquad M_{FB} = \frac{M_{FB} + 6\alpha_A M_{FA}}{(1 + 4\alpha_A + 4\alpha_B + 12\alpha_A\alpha_B)}$$

$$\alpha_A = \frac{EI}{LK_A} \qquad \alpha_B = \frac{EI}{LK_B}$$

The stiffness matrix in global coordinates of a beam member ($i$) with semi-rigid end connections is not only influenced by the beam's modulus of elasticity ($E$), moment of inertia ($I$), cross-sectional area ($A$), and length ($L$), but also by the angle ($\alpha$) between the global and local coordinate systems. The aforementioned factors are elaborated in detail. Additionally, to assess the impact of in-span gravity loads on the beam with semi-rigid connections, the calculation of the fixed-end force vector ($\{r_F\}$) in member coordinates is necessary. In order to determine the fixed-end moments ($M_{FA}$ and $M_{FB}$) and fixed-end shears ($V_{FA}$ and $V_{FB}$) for the member with fixed connections, the equilibrium of the member must be considered.

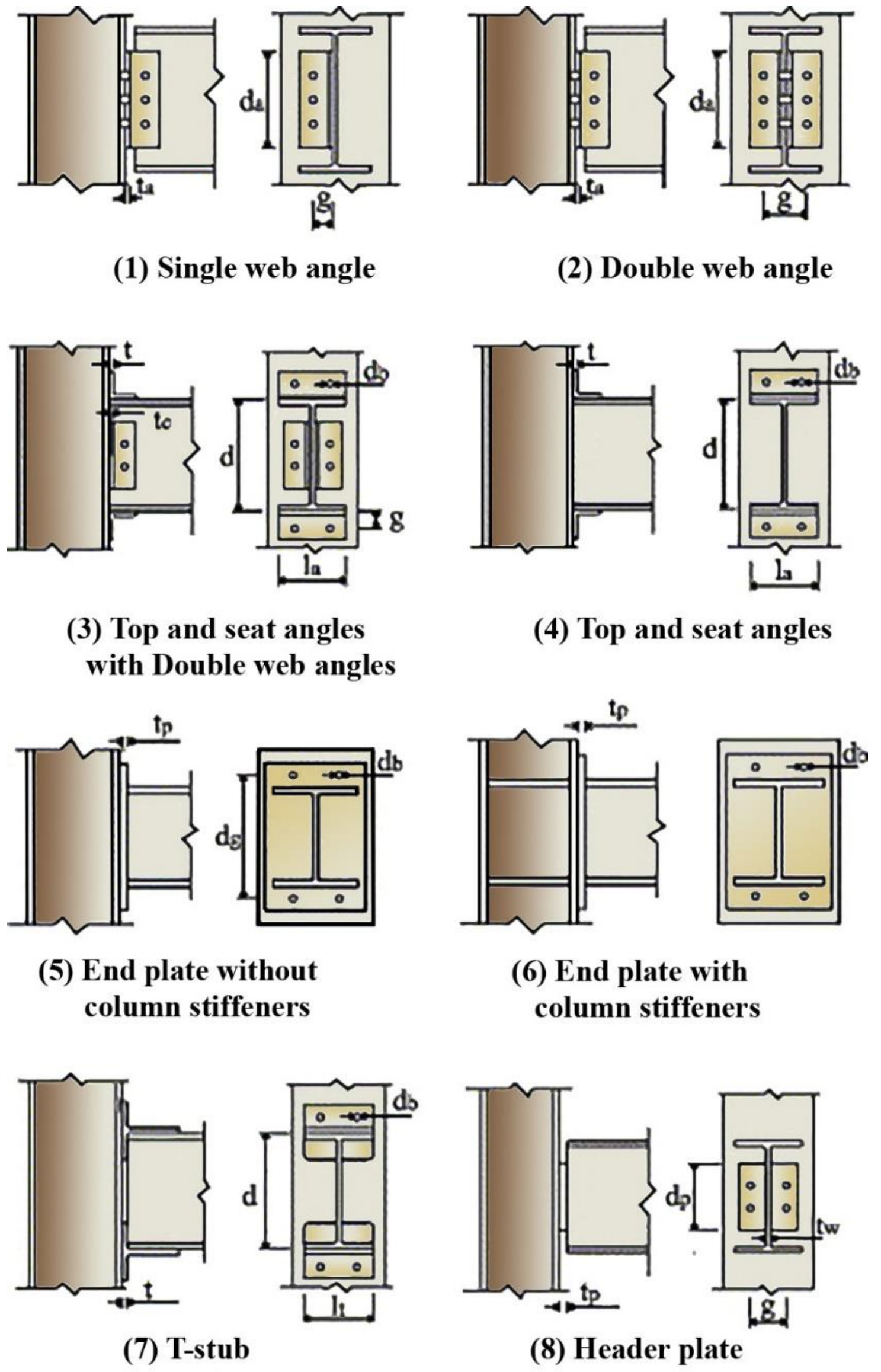

**Fig. 3** Intersection of membership functions

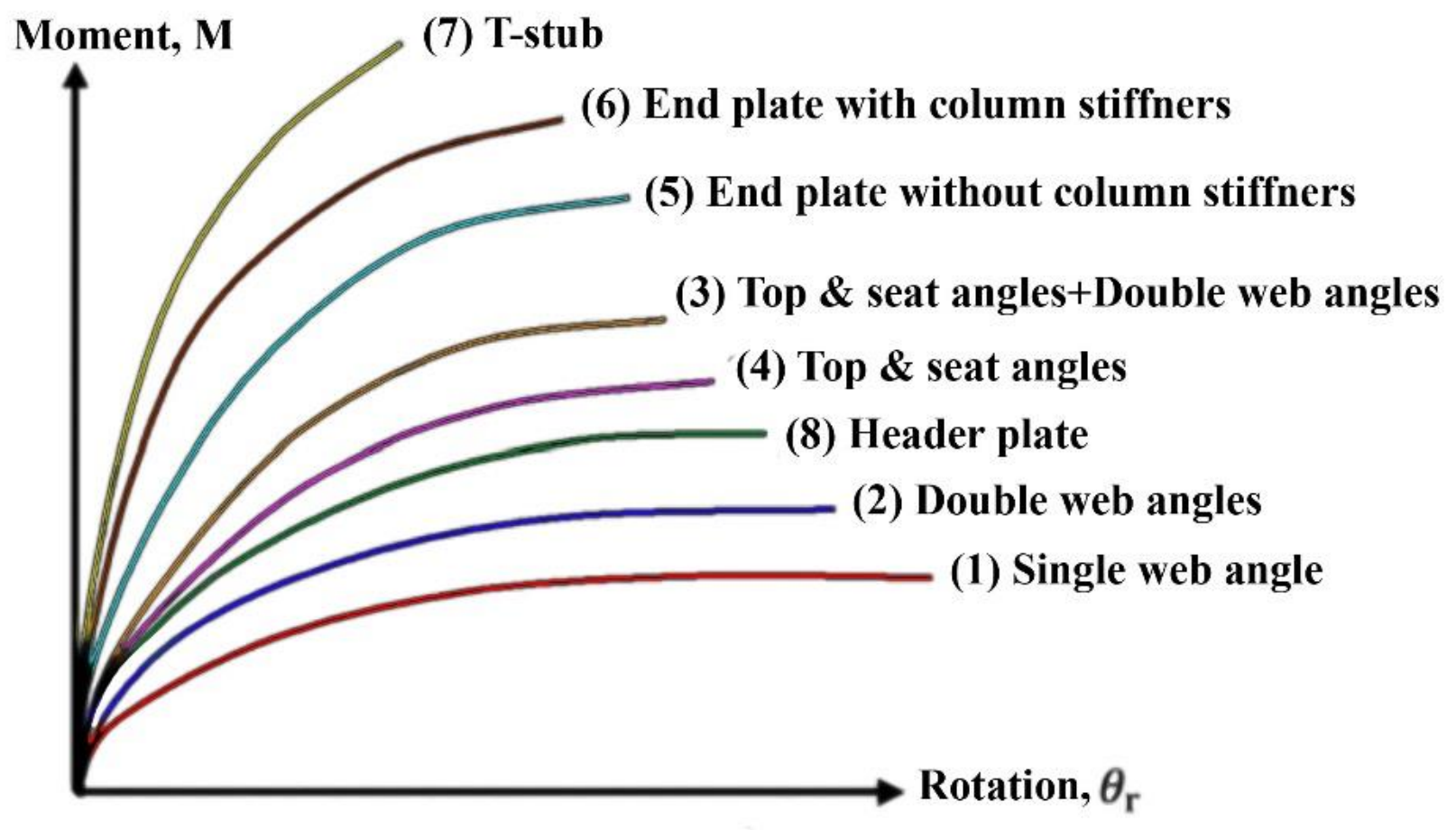

**Fig. 4** Moment-Rotation curves of semi-rigid connection types

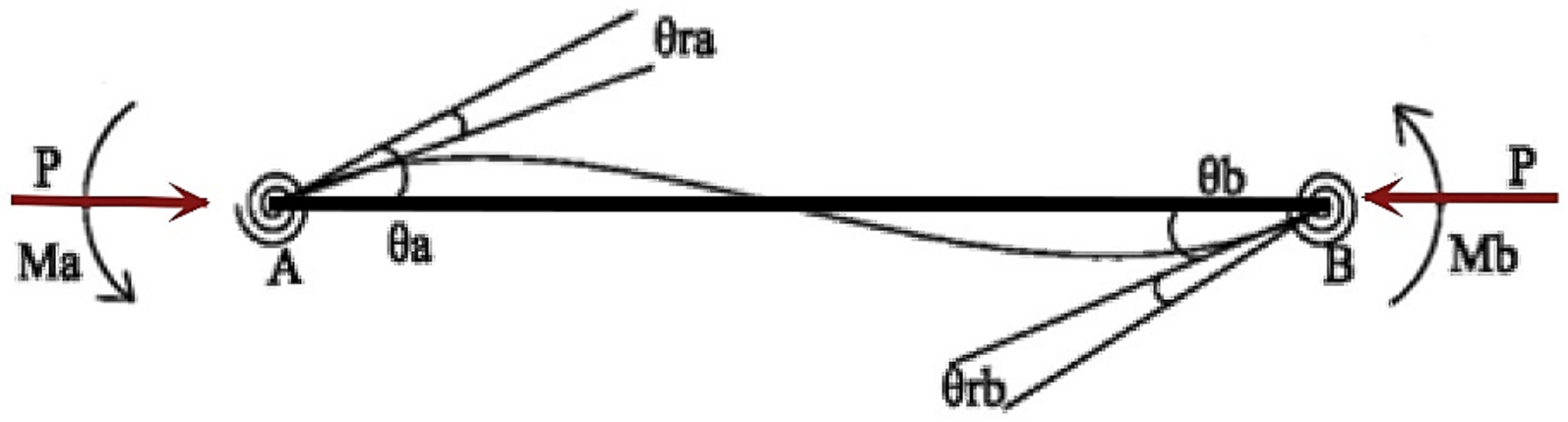

**Fig. 5** Intersection of membership functions

## 4. Results and discussion

For the purpose of weight optimization, three different examples were selected and compared, taking into consideration the use of both semi-rigid and rigid connections. All of the selected examples had a bay length of 5 meters and a story height of 3.2 meters. The elasticity modulus for all examples was consistent, being 2.1 GPa. The dead load and live load values were determined as 5886 N/m² and 1962 N/m², respectively, with the live load for the roof level being 1471.5 N/m². A standard set of steel sections, such as the American wide-flange sections ("W" Shape), were used as beams and column. The density of utilized steel was 77008 N/m². For seismic loads, Uniform Building Code (UBC, 1997) was employed. The code defines the seismic base shear as a design parameter.

$$V = C.W, \quad C = ABI/R \tag{11}$$

where $W$ and $C$ is the effective weight of frame, and the seismic response coefficient, respectively. $A$ is the Peak Ground Acceleration ($PGA$); $B$, $I$, and $R$ are the factors of spectral response, building importance, and response modification, respectively.

## 4.1. Verification of the design and method

The analysis of the frames in this study assumes a mutation possibility of 0.005 and a population size of 30, with modeling and analysis conducted using the OpenSees software. Table 1 presents the rotational stiffness values for various types of semi-rigid connections, as provided by Hayalioglu [31]. To confirm the accuracy of the semi-rigid frame connections in OpenSees, a beam with end springs subjected to uniform loading (Fig. 6) is analyzed and compared to findings of Abdul-Rassak Sultan [38] and Vatani Oskouei [39], with the results presented in Table 2. Furthermore, to establish the reliability of the GA in MATLAB, some examples are created and compared to the outcomes obtained by Hayalioglu [31], with the outcomes presented in Table 2 for comparison.

**Table 1.** The rotational stiffness of semi-rigid connections

| Connections | **#1** | **#4** | **#5** | **#7** |
|---|---|---|---|---|
| K (N·cm/rad) | $833 \times 10^7$ | $2766 \times 10^7$ | $3325 \times 10^7$ | $4434 \times 10^7$ |

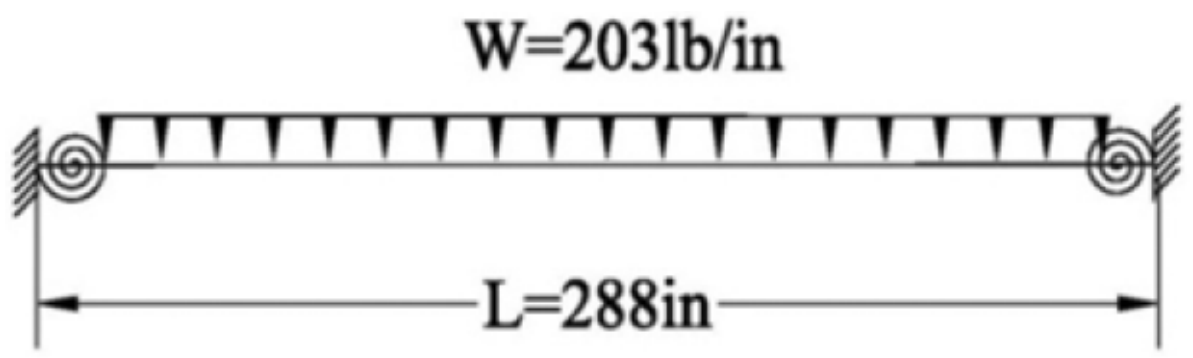

**Fig. 6** Beam with Rotational Spring

**Table 2.** Rotational stiffness values for semi-rigid connections

| Reference | Section | $E\ (k.in^2)$ | $I\ (in^4)$ | $Mu\ (k.in)$ | $Ki$ |
|---|---|---|---|---|---|
| [38] | W8×21 | 29000 | 75.3 | 406 | 27795 |
| [39] | W8×21 | 29000 | 75.3 | 405.53 | 27795 |
| Current study | W8×21 | 29000 | 75.3 | 405.852 | 27795 |
| | | Optimization result using GA | | | |
| | Frame example | | Weight (ton) | Top displacement (cm) | |
| [31] | 9-story, single-bay | | 11.794 | 7.87 | |
| Current study | 9-story, single-bay | | 11.745 | 7.96 | |

## 4.2. Three-story frame

In this study, a steel frame consisting of three bays and three stories was designed with both semi-rigid and rigid connections, and a linear analysis was carried out. The specification of the frame including the classified group labels is displayed in Fig. 7. Optimal results were obtained after 75 generations, and the process was repeated ten times to determine the frame's lightest weight. The results of structural weight for the frame with rigid and semi-rigid connections is illustrated in Fig. 8 throughout the optimization process. The results presented in Table 8 indicate that using semi-rigid connections resulted in a decrease in weight of 2.4-11.7% when compared to rigid connections. Additionally, Table 3 reveals that the fuzzy-GA led to a decrease in weight of 1.2-3.5% in comparison to the simple GA. Furthermore, the frame with semi-rigid connections exhibited an increase in lateral displacement of the roof by 4.2-6.5% in comparison with the frame with rigid connections.

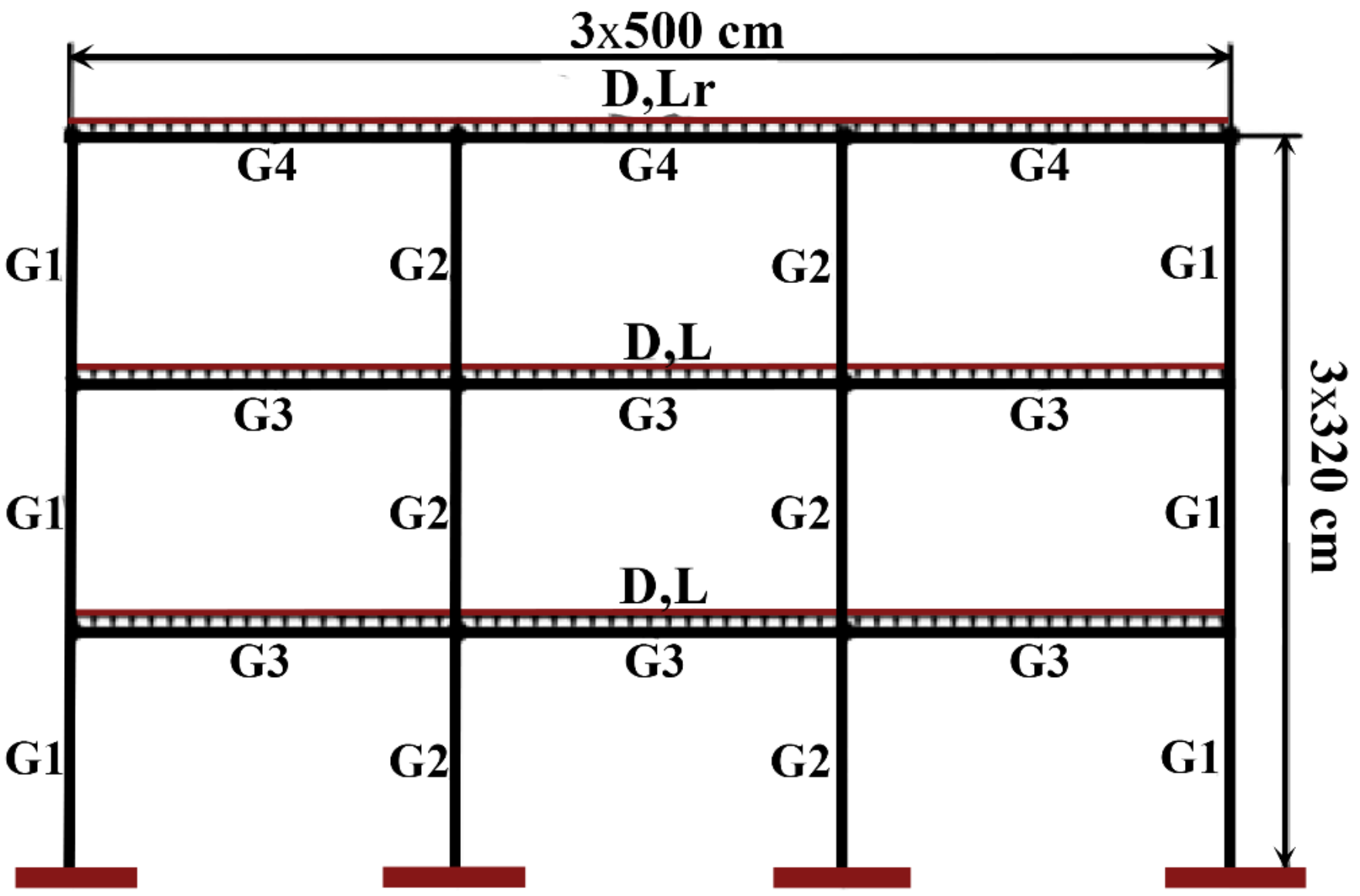

**Fig. 7** Three-bay three-story frame

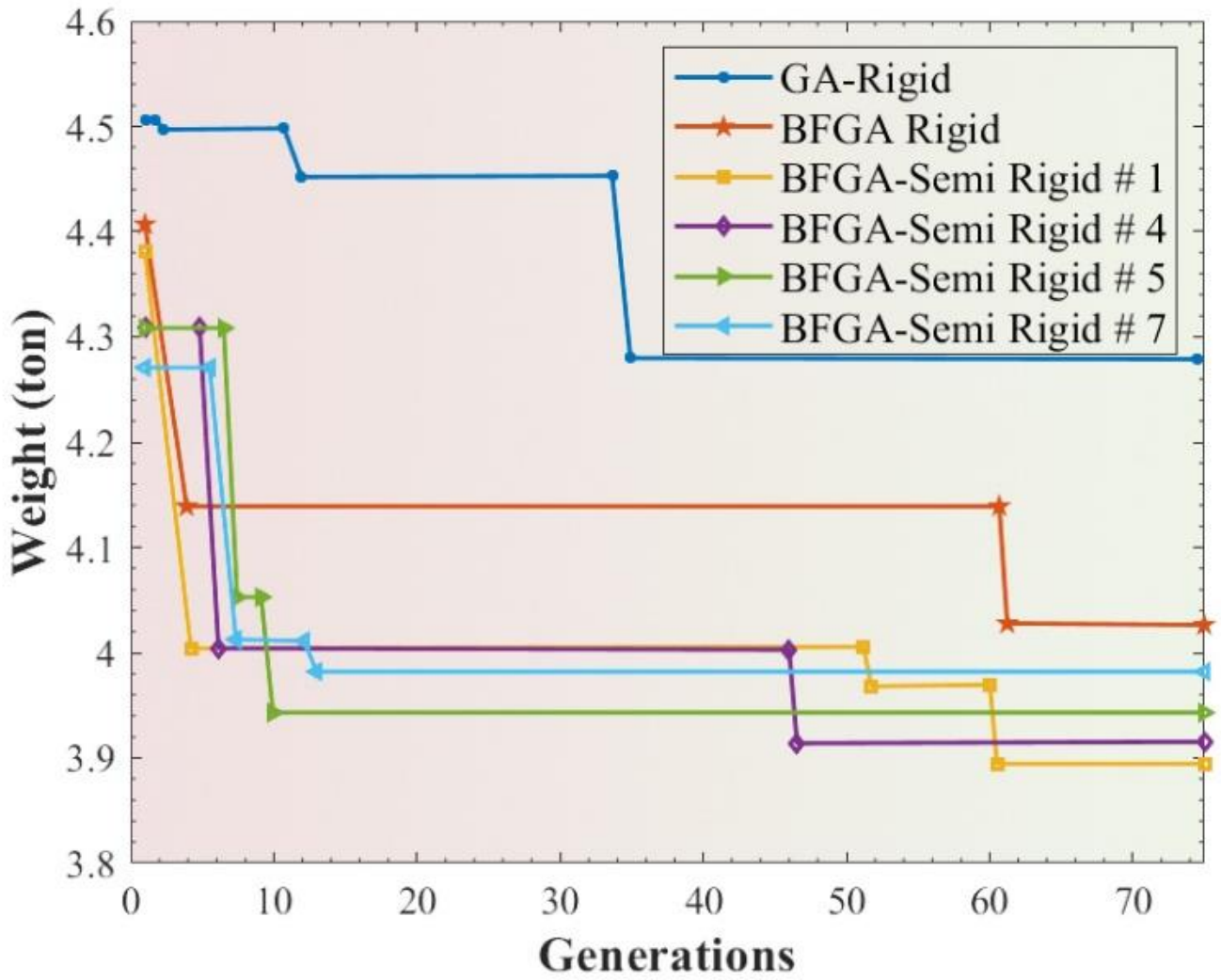

**Fig. 8** Convergence history of Three-bay three-story frame

**Table 3.** Optimum result comparison for the three-story frame

| Group no. | Semi-rigid connection type #1 | #4 | #5 | #7 | Rigid connection |
|---|---|---|---|---|---|
| | *Standard Genetic Algorithm* | | | | |
| G1 | $W16X26$ | $W16X26$ | $W16X26$ | $W16X40$ | $W12X16$ |
| G2 | $W16X31$ | $W16X31$ | $W14X34$ | $W14X53$ | $W16X40$ |
| G3 | $W16X36$ | $W16X36$ | $W14X38$ | $W16X36$ | $W14X43$ |
| G4 | $W16X26$ | $W14X30$ | $W16X26$ | $W14X30$ | $W14X34$ |
| Weight (Ton) | 3.8285 | 3.9174 | 4.0043 | 4.1780 | 4.2795 |
| Disp. (cm) | 2.6221 | 2.9294 | 2.8517 | 2.5728 | 2.4702 |
| | *Bilinear Fuzzy-Genetic Algorithm* | | | | |
| G1 | $W16X26$ | $W16X26$ | $W16X31$ | $W12X16$ | $W16X31$ |
| G2 | $W14X34$ | $W14X34$ | $W14X30$ | $W16X40$ | $W14X34$ |
| G3 | $W14X38$ | $W16X36$ | $W16X36$ | $W16X40$ | $W14X38$ |
| G4 | $W16X26$ | $W14X26$ | $W14X26$ | $W16X26$ | $W16X26$ |
| Weight (Ton) | 3.8925 | 3.9139 | 3.9431 | 3.9817 | 4.0269 |
| Disp. (cm) | 2.9517 | 2.8303 | 2.6887 | 2.5960 | 2.4210 |

## 4.3. Five-story frame

The optimum design of rigid and semi-rigid connections in a five-story steel frame with three bays was compared using linear analysis. The frame's topology, dimensions, and element labels were displayed in Fig. 9. The optimum results for simple fuzzy-GA and GA after 100 generations were tabulated in Table 4. To achieve the minimum weight of the frame, the procedure was repeated 10 times. Figure 10 illustrates the generational history of the optimized frame with both semi-rigid and rigid connections. After 55 generations, the fuzzy GA yielded almost the same minimum weight. Table 4 displays that the weight of the semi-rigid frame was 3.5-8.9% less than the rigid frame. In addition, Table 5 indicates that the fuzzy-GA produced a weight reduction of 4.7-6% in comparison with the simple GA. However, the lateral displacement of the roof in the semi-rigid frame increased by 9.2-22.4% when compared to the rigid frame. Table 1 presents the values of rotational stiffness.

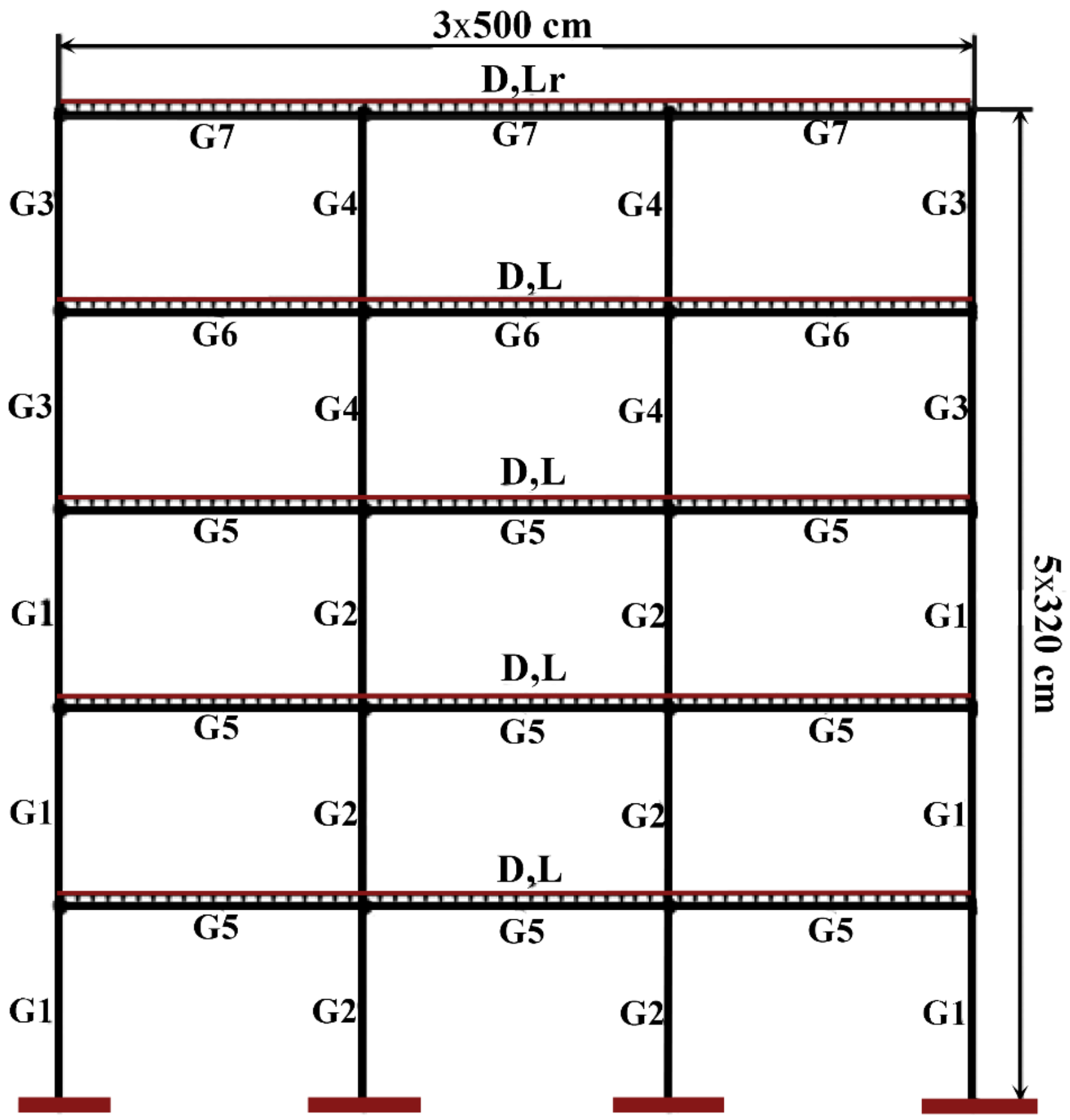

**Fig. 9** Three-bay five-story frame

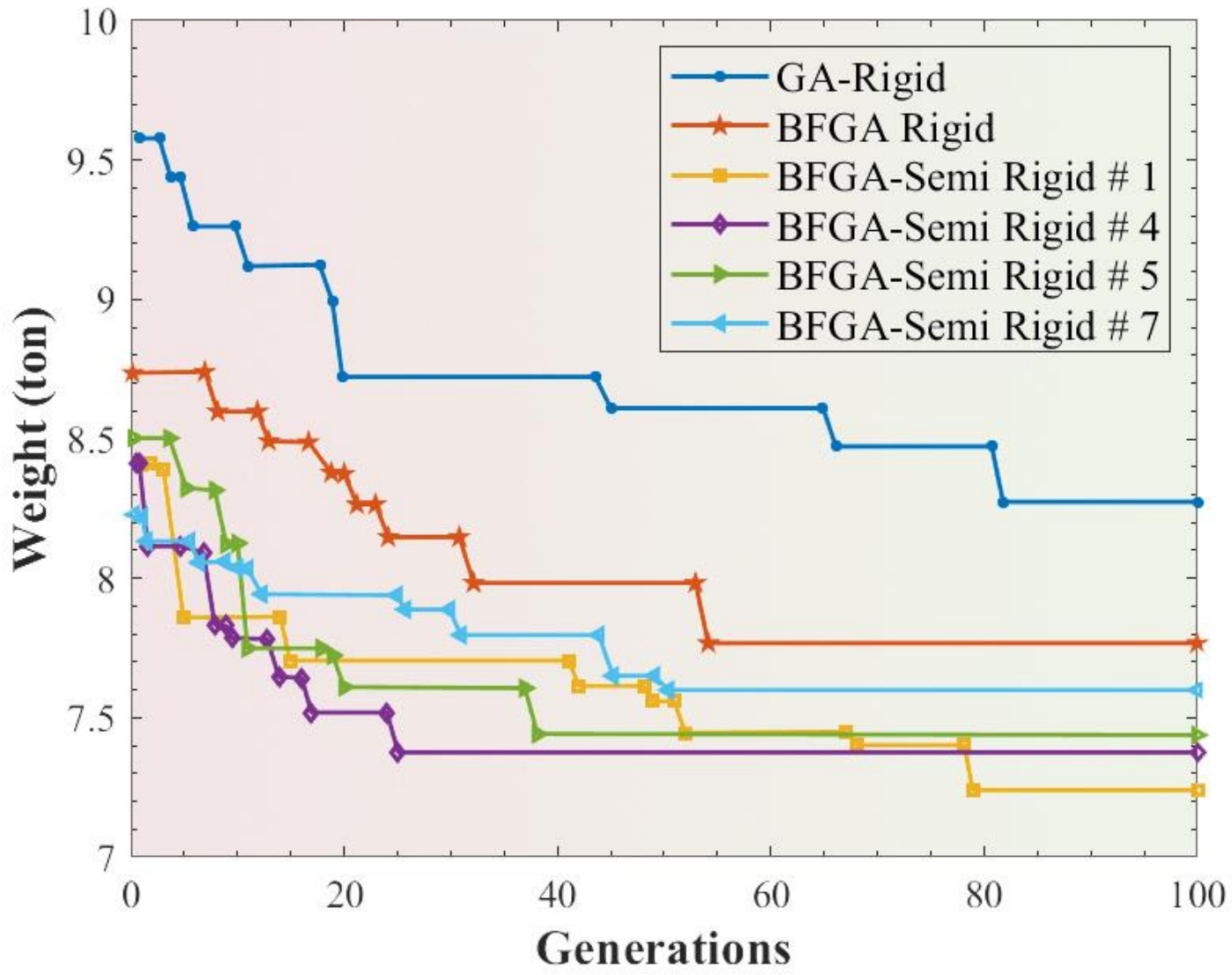

**Fig. 10** Convergence history of Three-bay five-story frame

**Table 4.** Optimum result comparison for the five-story frame

| Group no. | Semi-rigid connection type | | | | Rigid connection |
|---|---|---|---|---|---|
| | **#1** | **#4** | **#5** | **#7** | |
| | | | *Standard Genetic Algorithm* | | |
| G1 | W16X50 | W16X50 | W16X50 | W14X43 | W14X53 |
| G2 | W12X14 | W12X14 | W12X16 | W16X26 | W12X22 |
| G3 | W14X26 | W16X26 | W14X26 | W16X40 | W16X31 |
| G4 | W14X53 | W14X43 | W14X34 | W16X26 | W14X30 |
| G5 | W14X43 | W16X50 | W14X48 | W16X50 | W14X61 |
| G6 | W14X38 | W16X36 | W14X43 | W16X36 | W14X43 |
| G7 | W16X26 | W16X26 | W14X30 | W16X26 | W16X26 |
| Weight (Ton) | 7.5912 | 7.8874 | 7.9050 | 7.9866 | 8.2698 |
| Disp. (cm) | 6.9450 | 6.5710 | 6.3265 | 6.1993 | 5.6732 |
| | | | *Bilinear Fuzzy-Genetic Algorithm* | | |
| G1 | W16X50 | W16X50 | W14X48 | W16X40 | W16X40 |
| G2 | W12X16 | W12X16 | W12X16 | W16X26 | W16X26 |
| G3 | W16X31 | W14X34 | W14X34 | W16X40 | W14X38 |
| G4 | W14X43 | W14X43 | W16X40 | W14X30 | W16X26 |
| G5 | W14X43 | W16X40 | W16X45 | W16X45 | W14X48 |
| G6 | W14X34 | W14X34 | W14X34 | W14X34 | W14X38 |
| G7 | W16X26 | W16X26 | W16X26 | W14X26 | W16X26 |
| Weight (Ton) | 7.2482 | 7.3848 | 7.4434 | 7.6078 | 7.7755 |
| Disp. (cm) | 6.9959 | 6.7587 | 6.6899 | 6.2156 | 5.3740 |

## 4.4. Nine-story frame

A nine-story three-bay frame with rigid and semi-rigid connections is designed and subjected to linear analysis. The frame's shape, dimensions, and element labels are depicted in Figure 11. The optimal design results for the simple and fuzzy GA after 100 generations are presented in Tables 5 and 7, respectively. The process was iterated ten times to achieve the optimum frame weight.

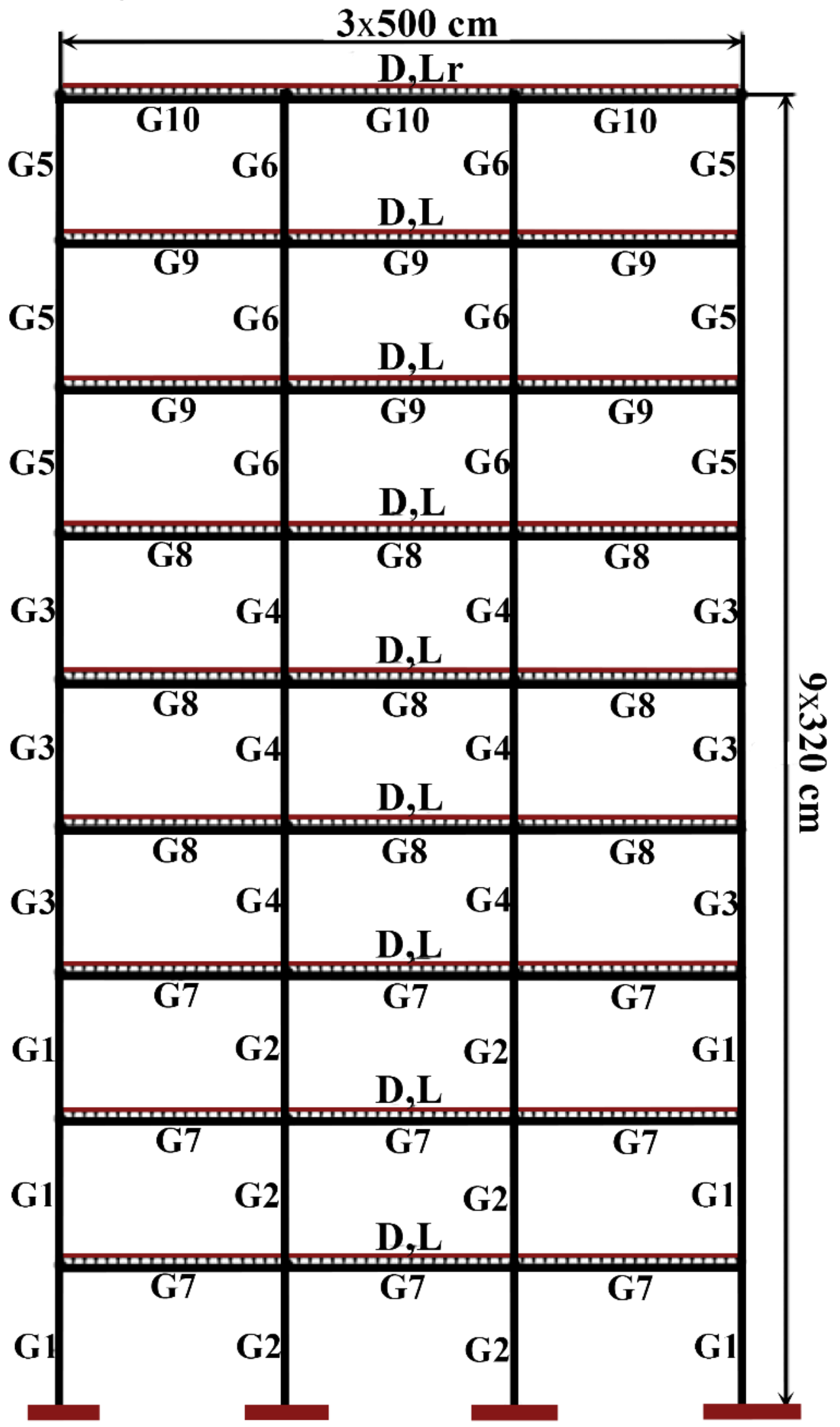

**Fig. 11** Three-bay nine-story frame

Figure 12 depict the weight variation for the semi-rigid and rigid frame, respectively, during the optimal design process across generations. The optimal design presented in Table 5 represents that the semi-rigid frame is lighter in weight by 0.8-5.2% than the rigid connection frame. Results show that the fuzzy-GA yields 1-1.5% decrease in weight in comparison with the standard GA. However, the semi-rigid frame experiences an increased lateral displacement of 14.5-19.3% compared to the rigid connection frame.

**Table 5.** Optimum result comparison for the nine-story frame

| Group no. | Semi-rigid connection type | | | | Rigid connection |
|---|---|---|---|---|---|
| | **#1** | **#4** | **#5** | **#7** | |
| | *Standard Genetic Algorithm* | | | | |
| G1 | $W24X55$ | $W24X55$ | $W24X55$ | $W24X55$ | $W24X55$ |
| G2 | $W24X55$ | $W21X62$ | $W27X102$ | $W21X55$ | $W21X55$ |
| G3 | $W18X60$ | $W18X60$ | $W18X60$ | $W18X71$ | $W24X62$ |
| G4 | $W21X55$ | $W21X55$ | $W24X55$ | $W18X60$ | $W24X55$ |
| G5 | $W18X86$ | $W18X71$ | $W24X55$ | $W21X55$ | $W21X55$ |
| G6 | $W21X55$ | $W21X55$ | $W24X55$ | $W24X55$ | $W21X55$ |
| G7 | $W18X46$ | $W18X50$ | $W18X50$ | $W14X68$ | $W18X50$ |
| G8 | $W18X46$ | $W18X50$ | $W18X50$ | $W18X50$ | $W14X68$ |
| G9 | $W18X46$ | $W18X46$ | $W18X40$ | $W18X46$ | $W14X48$ |
| G10 | $W12X30$ | $W18X46$ | $W18X46$ | $W18X50$ | $W12X30$ |
| Weight (ton) | 18.2057 | 18.8888 | 18.9883 | 19.0147 | 19.1564 |
| Disp. (cm) | 7.3313 | 7.2920 | 7.1709 | 7.0368 | 6.1413 |
| | *Bilinear Fuzzy-Genetic Algorithm* | | | | |
| G1 | $W21X55$ | $W21X55$ | $W21X55$ | $W24X55$ | $W21X55$ |
| G2 | $W21X55$ | $W21X55$ | $W24X55$ | $W24X55$ | $W24X55$ |
| G3 | $W24X62$ | $W24X62$ | $W24X55$ | $W24X55$ | $W24X62$ |
| G4 | $W24X62$ | $W21X62$ | $W24X55$ | $W21X55$ | $W21X55$ |
| G5 | $W24X55$ | $W24X55$ | $W24X55$ | $W21X68$ | $W21X55$ |
| G6 | $W24X62$ | $W21X55$ | $W21X55$ | $W21X101$ | $W21X73$ |
| G7 | $W18X50$ | $W18X46$ | $W18X46$ | $W18X46$ | $W18X50$ |
| G8 | $W18X46$ | $W18X50$ | $W24X55$ | $W18X50$ | $W18X50$ |
| G9 | $W18X40$ | $W18X46$ | $W18X46$ | $W18X46$ | $W18X40$ |
| G10 | $W18X76$ | $W14X74$ | $W12X30$ | $W12X30$ | $W18X46$ |
| Weight (ton) | 18.2209 | 18.4792 | 18.8210 | 18.8487 | 18.9703 |
| Disp. (cm) | 6.7156 | 6.5056 | 6.3493 | 6.0424 | 5.6339 |

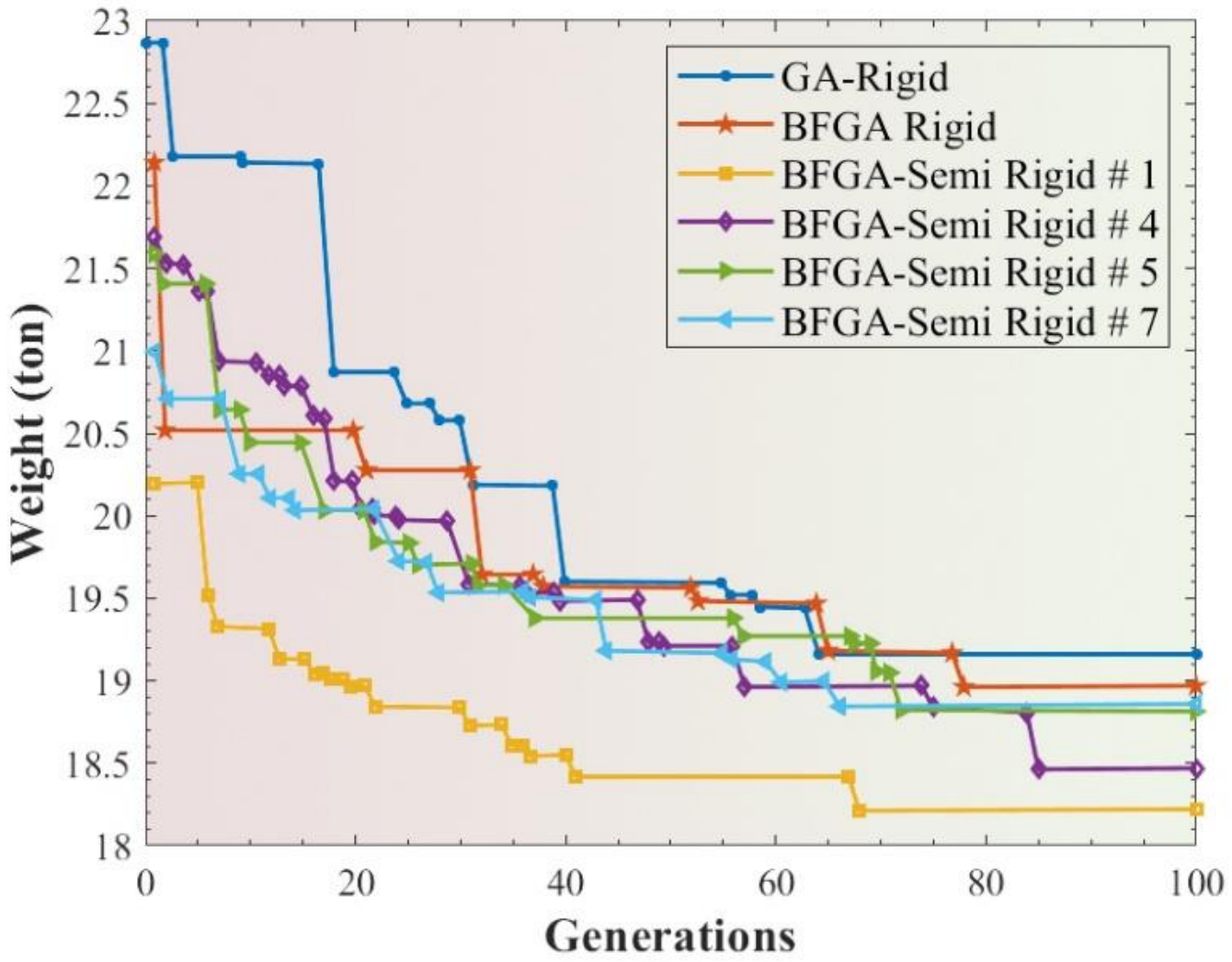

**Fig. 12** Convergence history of Three-bay nine-story frame

## 5. Conclusions

This chapter utilized a hybrid approach of fuzzy logic and genetic algorithm to optimize the structural weight of rigid and semi-rigid steel frames, while satisfying the displacement and stress constraints. To improve convergence speed, the genetic algorithm employs a bilinear membership function. Predefined scale factors are proposed to be used in the objective function to balance two terms of the equation and improve convergence. The displacement constraints have a greater effect on taller frames compared to other parameters, and the difference in weight between the optimized frames with semi-rigid and rigid connections is small for taller frames. Conclusions drawn from the design examples include:
The use of semi-rigid connections instead of rigid ones may cause the roof to displace laterally by a higher percentage, ranging from 4.2% to 22.4%. When using genetic algorithms to optimize steel frame design, incorporating fuzzy logic can lead to a reduction in the total weight of the structure by 1% to 6% compared to using simple genetic algorithms. Furthermore, the convergence rate can improve, and better solutions can be obtained with the use of fuzzy logic. The stiffness of the connections plays a key role in determining the lateral displacement and the overall structural weight. Increasing the stiffness of the connections can lead to a decrease in lateral displacement and an increase in the total weight. Conversely, using semi-rigid connections can reduce the weight of the frame by 0.8% to 11.7%. Parameters of the genetic algorithm, such as the possibility of mutation and crossover probability, affect convergence velocity; and the Population size affects the optimum weight, with larger population size resulting in a small decrease in weight but a large increase in computing time.